\documentclass[preprint,3p,times,twocolumn]{elsarticle}

\usepackage{amsmath,amssymb,amsfonts}
\usepackage{algorithmic}
\usepackage{graphicx}
\usepackage{textcomp}
\usepackage[linesnumbered,ruled]{algorithm2e}
\usepackage{bm}
\usepackage{xcolor}
\usepackage{array}
\usepackage{comment}
\usepackage[colorlinks,allcolors=blue]{hyperref} 
\usepackage{multirow}
\usepackage{soul}

\SetKwInput{KwInput}{Input}                
\SetKwInput{KwOutput}{Output}              
\SetKwInput{kwConstraint}{Constraint}      
\SetKwBlock{KwTA}{TA}{end}
\SetKwBlock{KwDH}{Feature Contributor}{end}
\SetKwBlock{KwLC}{Label Contributor}{end}
\SetKwBlock{KwCSP}{CSP}{end}
\SetKwFunction{PLSTrain}{PLS.train}
\SetKwFunction{PPPLSTrain}{P3LS.train}
\SetKwFunction{PPPLSRecover}{P3LS.recover}
\SetKwFunction{PPPLSEV}{P3LS.explained_variance}
\SetKwFunction{PPPLSPredict}{P3LS.predict}
\SetKwFunction{FedMPLSPredict}{FedMPLS.predict}
\SetKwProg{Fn}{Function}{:}{End}

\graphicspath{{figures/}}

\journal{Journal of Process Control}

\begin{document}

\begin{frontmatter}


\title{P3LS: Partial Least Squares under Privacy Preservation}
\author[inst1]{Nguyen Duy Du}
\author[inst1]{Ramin Nikzad-Langerodi\corref{cor}}
\ead{ramin.nikzad-langerodi@scch.at}
\cortext[cor]{Corresponding author}

\affiliation[inst1]{organization={Software Competence Center Hagenberg},
            country={Austria}}
            
\begin{abstract}

Modern manufacturing value chains require intelligent orchestration of processes across company borders in order to maximize profits while fostering social and environmental sustainability. However, the implementation of integrated, systems-level approaches for data-informed decision-making along value chains is currently hampered by privacy concerns associated with cross-organizational data exchange and integration. We here propose Privacy-Preserving Partial Least Squares (P3LS) regression, a novel federated learning technique that enables cross-organizational data integration and process modeling with privacy guarantees. P3LS involves a singular value decomposition (SVD) based PLS algorithm and employs removable, random masks generated by a trusted authority in order to protect the privacy of the data contributed by each data holder. We demonstrate the capability of P3LS to vertically integrate process data along a hypothetical value chain consisting of three parties and to improve the prediction performance on several process-related key performance indicators. Furthermore, we show the numerical equivalence of P3LS and PLS model components on simulated data and provide a thorough privacy analysis of the former. Moreover, we propose a mechanism for determining the relevance of the contributed data to the problem being addressed, thus creating a basis for quantifying the contribution of participants.
\end{abstract}

\begin{graphicalabstract}
\includegraphics[scale=0.47]{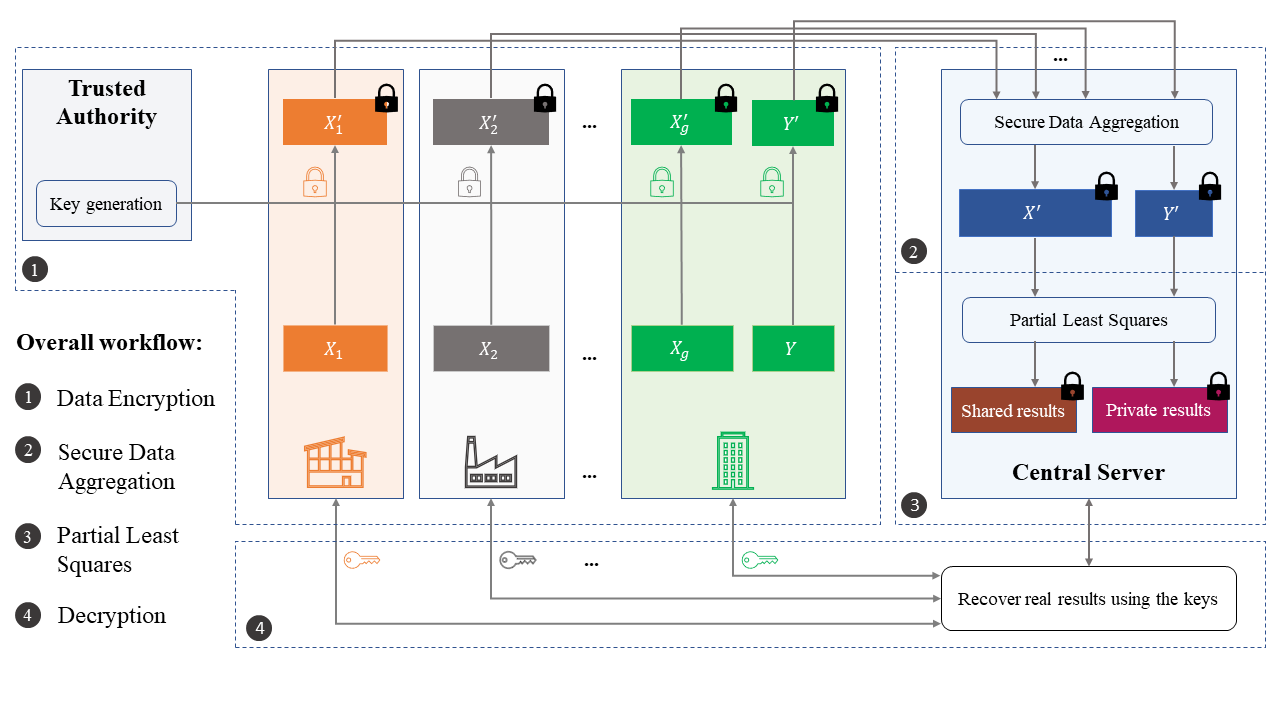}
\end{graphicalabstract}

\begin{highlights}
\item Highlight 1: We developed an approach that allows multiple companies along a value chain to collaboratively construct and exploit a cross-organizational process model while protecting their data privacy and security.
\item Highlight 2: We proposed an incentive mechanism that quantifies participants' contributions, thus creating a basis for a profit-sharing scheme and motivating the companies to join the federation and contribute high-quality data.
\item Highlight 3: We experimented on synthetic datasets to evaluate the effectiveness of the proposed method. The results show that our method outperformed the local model and yielded the same performance as the conventional approach of centralizing data.
\end{highlights}


\end{frontmatter}


\section{Introduction}
Value chains involving sequential manufacturing stages operated by different companies have become increasingly prevalent, e.g., in the automotive, semiconductor, pharmaceutical, and waste management industry \cite{colledani2014}. The transition away from (petroleum-based) virgin - and towards increasing use of recycled - feedstock further increases the mutual interaction and dependence of the involved processes. Thus, achieving consistent product characteristics requires closer collaboration of the stakeholders along value chains to be able to dynamically adapt the corresponding processes to increasingly variable input streams.

Distributed ledger technology (DLT) has become the technology of choice for trustworthy, cross-organizational information exchange and has been adopted in several industries such as manufacturing, automotive, real estate, energy, or healthcare \cite{lohmer2022}. DLT also forms the basis of various (non-) corporate digital passport initiatives, e.g., the European Blockchain Partnership (EBP). With the upcoming EU Battery regulation that calls for a mandatory digital battery passport by 2026, the pace for implementing digital passports is also picking up from the regulatory side \cite{ele2023}. Although digital passports have been promoted as the key enabler of the circular economy and successfully implemented to realize circular business models, they come with several limitations \cite{walden2021}. First and foremost, digital passports, whether DLT-based and equipped with encryption technology to protect sensitive data or not, always contain only those pieces of information that are \emph{a priori} deemed relevant for process and value chain optimization. However, it is not always clear which material/process/product specifications have an impact on downstream processes, and thus, it can be challenging to determine which information should be included in digital passports to ensure their maximum effectiveness in supporting a circular economy and sustainable development goals \cite{walden2021}.

With the ever-increasing complexity of industrial processes and amounts of data produced along value chains, AI and big data analytics have become obvious tools to help reveal such interactions across company borders. Joint analysis of (process) data can provide stakeholders with a more comprehensive view of the entire value chain, including the relationships between different process stages and the factors that impact the sustainability and circularity of the system as a whole. However, despite the overwhelming consensus that cross-organizational information exchange is imperative for increasing efficiency, resilience as well as social and environmental sustainability of the processes involved along value chains, privacy concerns related to data sharing still prevail among the relevant stakeholders.

Federated learning (FL) is a relatively new machine learning paradigm, first proposed by researchers at Google in 2016, that addresses privacy and security concerns associated with centralized data storage and processing. FL allows multiple parties to jointly train data-driven models while preserving data privacy – either by entirely avoiding data exchange or by means of data encryption \cite{du2023}. Vertical Federated Learning (VFL) refers to scenarios where the private datasets share the same sample space but differ in the feature space. Cross-organizational processes naturally fit this scenario because even though local datasets can be mapped using the product or batch IDs, the local features are often distinct because each company usually operates a different type of sub-process. Although some research efforts have been devoted to adopting VFL in the field of process modeling, limited progress has been made. In  \cite{hartebrodt2021} \cite{grammenos2020}, the authors have proposed federated variants of Principal Component Analysis (PCA), which is a widely used method for fault detection and diagnosis. Nevertheless, in these studies, the application of PCA to process monitoring was not investigated. In \cite{du2022}, Du \emph{et al.} recently proposed a multiblock principal component analysis (MPCA) based federated multivariate statistical process control (FedMSPC) approach to jointly model a semiconductor fabrication process over multiple stages that are operated by different companies. Not only was the joint model more efficient (compared to the model trained on the downstream process only) for predicting process faults, but it also revealed important interactions between the two process stages that led to faulty batches in the downstream process. This study is among the first to demonstrate the prospects of (vertical) federated data analysis and process modeling for collaborative risk identification, decision-making, and value chain optimization.

However, as an unsupervised technique, FedMSPC's application is limited to scenarios where the relationship between process variables and some specific response variables (e.g., product quality, KPIs, etc.) is not of primary interest. Furthermore, previous methodology lack mechanisms to incentivize data federation, which is pivotal to fostering collaboration among stakeholders along the value chain. In particular, participants should be rewarded based on their contribution to the overall outcomes (e.g., cost savings) resulting from data federation. Accessing contribution based merely on data quantity is certainly not enough, as one party may contribute a large volume of data that doesn’t help much in solving the problem. Recent studies have proposed the use of the Shapley value in evaluating data contribution \cite{wang2019}\cite{wang2020}. However, since the Shapley value has an exponential time complexity, this approach poses significant inefficiency \cite{yang2023}.

To overcome the crucial shortcomings of MPCA-based MPSC, we here propose Privacy-preserving Partial Least Squares (P3LS). P3LS is a federated variant of Partial Least Squares (PLS), a popular supervised technique for process modeling \cite{macgregor1995}\cite{nomikos1995}. P3LS leverages Federated Singular Vector Decomposition (FedSVD) \cite{chai2022} to enable different companies to jointly train a PLS model while protecting private datasets. To create a basis for incentive mechanisms, we also propose a method for securely quantifying the contribution of participants based on the variance explained by the contributed data. 

In the next section, we first cover the background of the proposed method, including the concepts of PLS, FL, and FedSVD. In Section \ref{sec:p3ls}, the methodology and implementation of P3LS are described in detail. We then provide a security analysis in Section \ref{sec:security_analysis} and discuss potential real-world applications of P3LS in manufacturing in Section \ref{sec:applications}. After that, in Section \ref{sec:experiment}, we provide an empirical evaluation of P3LS on simulated data when its performance is compared with local and centralized PLS models. Finally, conclusions drawn from the results and potential topics for future research and development are presented in Section \ref{sec:discussion}.

\section{Background} \label{sec:background}

\subsection{Preliminaries}
We use boldfaced lower-case letters, e.g., $\bm{x}$, to denote vectors and boldfaced upper-case letters, e.g., $\bm{X}$, to denote matrices. The subscript is associated with indices of different parties, e.g., the matrix $\bm{X}_i$  denotes the $i^{th}$ party's data.



\subsection{Partial Least Squares}
Suppose the process data of $m$ samples (e.g., batches) and $n$ variables is denoted as $\bm{X} \in \mathbb{R}^{m \times n}$, and the corresponding response variables (e.g., critical quality attributes of the final product) is stored in $\bm{Y} \in \mathbb{R}^{m \times l}$. The general underlying model of PLS applied on $\bm{X}$ and $\bm{Y}$ is:
\begin{align*}
   \bm{X} &= \bm{T}\bm{P}^\top + \bm{\Theta}\\
   \bm{Y} &= \bm{U}\bm{Q}^\top + \bm{\Phi},
\end{align*}

where 
\begin{itemize}
    \item $\bm{T} \in \mathbb{R}^{m \times k}$ and $\bm{U} \in \mathbb{R}^{m \times k}$ are, respectively, the $\bm{X}$ scores and the $\bm{Y}$ scores.
    \item $k$ is the number of latent variables (LVs), often determined by means of cross-validation \cite{abdi2003}.
    \item $\bm{P} \in \mathbb{R}^{n \times k}$ and $\bm{Q} \in \mathbb{R}^{l \times k}$ are, respectively, $\bm{X}$ loadings and $\bm{Y}$ loadings.
    \item $\bm{\Theta} \in \mathbb{R}^{m \times n}$ and $\bm{\Phi} \in \mathbb{R}^{m \times l}$ are the residuals matrices.
\end{itemize}
 
The decompositions of $\bm{X}$ and $\bm{Y}$ are made in a way that maximizes the covariance between $\bm{T}$ and $\bm{U}$. Different implementations of PLS exist, among which the most popular are NIPALS \cite{geladi1986}, SIMPLS \cite{de1993}, and Singular Vector Decomposition (SVD) \cite{wehrens2007}\cite{le2008}. In this study, we propose a privacy-preserving variant of PLS that leverages a federated version of SVD. A description of SVD-based PLS is provided in Algorithm \ref{alg:pls_training}.

\begin{algorithm*}[t]
\DontPrintSemicolon
  
  \KwInput{$\bm{X}$, $\bm{Y}$}
  \KwOutput{$\bm{T}$, $\bm{Q}$, $\bm{W}$, $\bm{P}$, $\bm{B}$, $\bm{R}$}
  \Fn{\PLSTrain{$\bm{X}$, $\bm{Y}, k$}}{
            
        Initialize two matrices: $\bm{E} := \bm{X}$, and $\bm{F} := \bm{Y}$\\
        \For{$j = 1 \rightarrow k$}{
            Calculate the cross-product matrix $\bm{S}$: $\bm{S} = \bm{E}^\top\bm{F}$.\\
            Perform SVD on $\bm{S}$ and extract the first left and right singular vectors, $\bm{w}$ and $\bm{v}$.\\
            Calculate scores $\bm{t}$ and $\bm{u}$:
            \begin{equation*}
                \bm{t} = \bm{E}\bm{w}, \quad \bm{u} = \bm{F}\bm{v}
            \end{equation*}
        
            Obtain loadings by regressing against the same vector $\bm{t}$:
            \begin{equation*}
                \bm{p} = \frac{\bm{E}^\top\bm{t}}{\bm{t}^\top\bm{t}}, \quad \bm{q} = \frac{\bm{F}^\top\bm{t}}{\bm{t}^\top\bm{t}}
            \end{equation*}

            Deflate the data matrices $\bm{E}$ and $\bm{F}$:
            \begin{equation*}
                \bm{E} = \bm{E} - \bm{t}\bm{p}^\top, \quad \bm{F} = \bm{F} - \bm{t}\bm{q}^\top
            \end{equation*}

            Save $\bm{w}$, $\bm{t}$, $\bm{p}$, and $\bm{q}$.
        }
        Use vectors $\bm{w}$, $\bm{t}$, $\bm{p}$ and $\bm{q}$ as columns to form matrices $\bm{W}$, $\bm{T}$, $\bm{P}$ and $\bm{Q}$, respectively.\\
        Compute the rotation matrix (denoted as $\bm{R}$) and regression coefficients (denoted as $\bm{B}$):
            \begin{equation*}
                \bm{R} = \bm{W}(\bm{P}^\top\bm{W})^{-1}, \quad \bm{B} = \bm{R}\bm{Q}^\top
            \end{equation*}
  }
\caption{PLS Training}\label{alg:pls_training}
\end{algorithm*}

After the model is trained, it can be used to calculate scores and predict the output for unseen data $\bm{X}_{new}$:
\begin{align*}
    \bm{T}_{new} &= \bm{X}_{new}\bm{R}\\
    \hat{\bm{Y}}_{new} &= \bm{X}_{new}\bm{B}
\end{align*}


The PLS algorithms commonly require centralized data, which poses challenges in cross-organizational processes where each company possesses data only from its own production line. Privacy and security concerns further restrict the sharing of data among companies. While some variations of PLS have aimed to handle multi-source data \cite{macgregor1994}\cite{wangen1989}, they still assume full data accessibility, failing to address the critical issue of data privacy.

\subsection{Federated Learning}

\begin{figure}[th]
\includegraphics[scale=0.4]{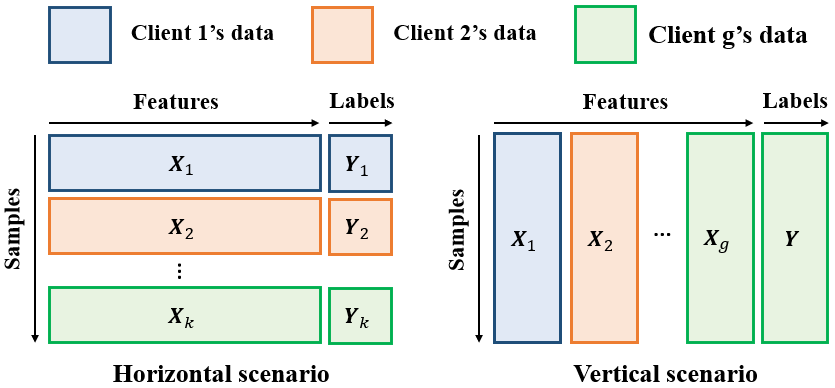}
\centering
\caption{Federated Learning categorization.}
  \label{fig:fl_cat}
\end{figure}

FL is a collaborative learning paradigm that allows multiple clients to jointly train a global machine learning model without revealing private data \cite{mcmahan2017}. Depending on how data is distributed among the clients, FL can be generally categorized into two settings, namely Horizontal Federated Learning (HFL) and Vertical Federated Learning (VFL). HFL refers to scenarios where private datasets share the same feature space but differ in the sample space. Conversely, VFL consists of situations where private datasets share the same sample space but differ in the feature space. The difference between the two scenarios is demonstrated in \textbf{Figure \ref{fig:fl_cat}}. As noted by \cite{li2021}\cite{yang2023}\cite{khan2022}, most of the current research in the field of privacy-preserving machine learning concentrates on the horizontal schema, while the vertical setting has received inadequate devotion.

\subsection{Federated Singular Vector Decomposition}


FedSVD, proposed in \cite{chai2022}, is a lossless approach for securely conducting SVD on distributed data without data centralization.

Suppose $\bm{H} \in \mathbb{R}^{m \times m}$ and $\bm{G} \in \mathbb{R}^{m \times m}$ are two random, orthogonal matrices. For any matrix $\bm{X} \in \mathbb{R}^{m \times n}$, it holds that $\bm{X}' = \bm{H}^\top\bm{X}\bm{G}$ has the same singular values as $\bm{X}$, and their singular vectors can be transformed into each other by a linear transformation, i.e.,
\begin{align}
\begin{aligned}\label{eq:fedsvd}
    \bm{w} &= \bm{H}\bm{w}'\\
    \bm{v} &= \bm{G}\bm{v}',
\end{aligned}
\end{align}
with $\bm{w}$ and $\bm{v}$, and $\bm{w}'$ and $\bm{v}'$ denoting the first left and right singular vectors of $\bm{X}$ and $\bm{X}'$, respectively.

In order to operate, FedSVD requires a trusted authority (TA) to handle the key generation (i.e., matrices $\bm{H}$ and $\bm{G}$) and a computation service provider (CSP) to perform data aggregation and matrix decomposition. The detailed descriptions of the method can be found in \cite{chai2022}. Briefly, FedSVD has four main steps:
\begin{itemize}
    \item \textbf{Step 1}: The TA generates random orthogonal matrices and creates shared and private keys.
    \item \textbf{Step 2}: Each data holder downloads its proprietary key and masks the private data.
    \item \textbf{Step 3}: The CSP aggregates the masked data using a secure aggregation protocol, performs SVD, and obtains masked singular vectors.
    \item \textbf{Step 4}: The data holders collaborate with the CSP to recover the real results using its given keys.
\end{itemize}

The approach was proved to be confidential under the semi-honest setting where the TA is fully trusted while the CSP and data holders are semi-honest \cite{chai2022}. More specifically, it assumes that the CSP and data holders will strictly follow the pre-specified protocol, even though they may try to deduce the private knowledge of others. In addition, The security protocol also requires the absence of collusion among the participants. Collusion denotes any secret or illicit agreement between entities aiming to compromise the system's security. For instance, collusion might occur if the TA or a data holder shares encryption keys with the CSP, allowing the decryption of masked data or masked model components, thereby undermining security.

\section{Privacy-Preserving Partial Least Squares} \label{sec:p3ls}


\subsection{Participants and roles}

The hypothetical scenario involves $g$ data holders, each holding a partition of features of the same samples. Let $\bm{X}_i \in \mathbb{R}^{m \times n_i}$ denotes the features owned by the $i^{th}$ data holder, where $m$ is the number of samples (batches) and $n_i$ is the number of private features. And let $\bm{Y} \in \mathbb{R}^{m \times l}$ represent the labels of the $m$ samples. Without loss of generality, we assume that $\bm{Y}$ is privately owned by the $g^{th}$ data holder.

Based on their contribution, we categorize data holders into two types: Feature Contributor (FC) and Label Contributor (LC). A data holder that contributes only a feature block (i.e., $\bm{X}_i$) is designated as a Feature Contributor. On the other hand, a data holder that contributes to the target block (i.e., $\bm{Y}$) is referred to as a Label Contributor. A feature contributor may also play the role of a label contributor. While there can be multiple label contributors, the present work focuses on scenarios where there is only one single $\bm{Y}$ block from one label contributor. An extension to handle multi-label-contributor applications will be investigated in future work.

Besides data holders, P3LS also requires the presence of a trusted authority and a central service provider. Their functions within the system closely mirror the roles they play in FedSVD.


\subsection{Overall workflow}

Suppose the data holders in the data federation aim at training a PLS model on the concatenated matrix $\bm{X} = [\bm{X}_1, \bm{X}_2, ..., \bm{X}_g]$ and the matrix $\bm{Y}$, where $\bm{X} \in \mathbb{R}^{m \times n}$ and $n = \sum_{i=1}^{g} n_{i}$. Assume that the matrices have been standardized to have zero means and unit variance. In this case, the P3LS decomposition results are the following:

\begin{align}
    [\bm{X}_1, ..., \bm{X}_g] &= \bm{T}[\bm{P}_1^\top, ..., \bm{P}_g^\top] + [\bm{\Theta}_1, ..., \bm{\Theta}_g]\\
    \bm{Y} &= \bm{U}\bm{Q}^\top + \bm{\Phi}
\end{align}


Accordingly, FC-$i$ gets $\bm{X}_i = \bm{T}\bm{P}^\top_i + \bm{\Theta}_i$, where $\bm{T}$ is a shared result among all participants, and $\bm{P}^\top_i \in \mathbb{R}^{n_i \times k}$ and $\bm{\Theta}_i$ are the private results. Meanwhile, since the LC is the sole contributor of the $\bm{Y}$-block, it must be the only party who has access to $\bm{U}$, $\bm{Q}$, and $\bm{\Phi}$. Additionally, in order to make predictions for new samples, all FCs must also know their local coefficients (i.e., $\bm{B}_i$, see Algorithm \ref{alg:pls_training}). Table \ref{table:fed_model_properties} summarizes the shared and private components of the P3LS model. It is strictly required that FC-$i$’s contributed data and private model components cannot be leaked to any other parties during the computation.

\begin{table}[hbt!]
\centering
\caption{Shared and private components of the P3LS model.}
\label{table:fed_model_properties}
\begin{tabular}{|c|c|l|}
\hline
\textbf{Role}         & \multicolumn{1}{l|}{\begin{tabular}[c]{@{}c@{}}\textbf{Shared}\\ \textbf{components}\end{tabular}} & \textbf{Private components}                                             \\ \hline
FC-$i$ & \multirow{2}{*}{$\bm{T}$ (X-scores)}            & \begin{tabular}[c]{@{}l@{}}$\bm{X}_i$ (Local features),\\ $\bm{P}_i$ (Local X-loadings),\\ $\bm{B}_i$ (Local coefficients),\\  $\bm{\Theta}_i$ (Local X-residuals)  \end{tabular} \\ \cline{1-1} \cline{3-3} 
LC     &                                                 & \begin{tabular}[c]{@{}l@{}}$\bm{U}$ (Y-scores),\\ $\bm{Q}$ (Y-loadings),\\ $\bm{Y}$ (Target block),\\ $\bm{\Phi}$ (Y-residuals)\end{tabular} \\ \hline
\end{tabular}
\end{table}

Overall, P3LS has the following three steps:
\begin{itemize}
    \item \textbf{Secure data aggregation}: The TA generates common and private keys, which are random orthogonal matrices. Each data holder downloads its proprietary key, masks their local data using the key, and then forwards the encrypted data to the CSP.
    \item \textbf{Sequential matrix decomposition}: The CSP aggregates the encrypted data, performs the standard SVD-based PLS algorithm, and obtains all encrypted model components.
    \item \textbf{Model decryption}: The data holders collaborate with the TA and the CSP to recover the real model components using both shared and private keys.
\end{itemize}

The pseudo-code of P3LS is shown in Algorithm \ref{alg:fedpls_training}. The following subsections will elaborate on the specifics of each step.

\begin{algorithm*}
\DontPrintSemicolon
  
  \KwInput{$\bm{X} = [\bm{X}_1,\ldots, \bm{X}_g]$, $\bm{Y}$}
  \KwOutput{$\bm{T}$, $\bm{Q}$, $\bm{W} = [\bm{W}_{1},\ldots, \bm{W}_{g}]$, $\bm{P} = [\bm{P}_{1},\ldots, \bm{P}_{g}]$, $\bm{B} = [\bm{B}_{1},\ldots, \bm{B}_{g}]$} 
  \Fn{\PPPLSTrain{$[\bm{X}_1,\ldots, \bm{X}_g]$, $\bm{Y}, k$}}{
         \KwTA(do:){
            Generate orthogonal matrices $\bm{A} \in \mathbb{R}^{m \times m}$, $\bm{H} \in \mathbb{R}^{n \times n}$, $\bm{G} \in \mathbb{R}^{l \times l}$. \\
            Split $\bm{H}^\top$ into $[\bm{H}_{1}^{\top}, \ldots, \bm{H}_{g}^{\top}]$ where $\bm{H}_i^\top \in \mathbb{R}^{n \times n_{i}}$.
         }
         \KwDH(do:){
           \For{$i = 1 \rightarrow g$, Feature Contributor $i$}{
                Download $\bm{A}$, $\bm{H}^\top_i$ from TA and compute
                $\bm{X}_{i}' = \bm{A}\bm{X}_{i}\bm{H}_{i}$\\
                Send $\bm{X}_{i}'$ to the CSP
           }
         }
         \KwLC(do:){
           Download $\bm{G}$ from the TA and computes $\bm{Y}' = \bm{A}\bm{Y}\bm{G}$\\
           Send $\bm{Y}'$ to the CSP
         }
        \KwCSP(do:){
            Receive $\bm{X}_{i}'$ and $\bm{Y}'$, then aggregate $\bm{X}'$ :
            \begin{equation*}
                \bm{X}' = \sum_{i=1}^{g} \bm{X}_{i}'
            \end{equation*}
            
            Assign $\bm{X}'$ and $\bm{Y}'$ to $\bm{E}'$ and $\bm{F}'$ respectively: $\bm{E}' := \bm{X}'$, and $\bm{F}' := \bm{Y}'$\\
            \For{$j = 1 \rightarrow k$}{
                Calculate the cross-product matrix $\bm{S}'$: $\bm{S}' = \bm{E}'^\top\bm{F}'$.\\
                Perform SVD on $\bm{S}'$ and extract the first left and right singular vectors, $\bm{w}'$ and $\bm{v}'$.\\
                Calculate scores $\bm{t}'$ and $\bm{u}'$:
                \begin{equation*}
                    \bm{t}' = \bm{E}'\bm{w}', \quad \bm{u}' = \bm{F}'\bm{v}'
                \end{equation*}
            
                Obtain loadings by regressing against the same vector $\bm{t}'$:
                \begin{equation*}
                    \bm{p}' = \frac{\bm{E}'^\top\bm{t}'}{\bm{t}'^\top\bm{t}'}, \quad \bm{q}' = \frac{\bm{F}'^\top\bm{t}'}{\bm{t}'^\top\bm{t}'}
                \end{equation*}

                Deflate the data matrices $\bm{E}'$ and $\bm{F}'$:
                \begin{equation*}
                    \bm{E}' = \bm{E}' - \bm{t}'\bm{p}'^\top, \quad \bm{F}' = \bm{F}' - \bm{t}'\bm{q}'^\top
                \end{equation*}

                Save $\bm{w}'$, $\bm{t}'$, $\bm{p}'$, and $\bm{q}'$.
            }
            Use vectors $\bm{w}'$, $\bm{t}'$, $\bm{p}'$ and $\bm{q}'$ as columns to form matrices $\bm{W'}$, $\bm{T}'$, $\bm{P}'$ and $\bm{Q}'$, respectively.\\
                Compute the masked rotation matrix and regression coefficients: %
                \begin{align*}
                    \bm{R}' &= \bm{W}'(\bm{P}'^\top\bm{W}')^{-1}\\
                    \bm{B}' &= \bm{R}'\bm{Q}'^\top
                \end{align*}
        }
        FCs and the LC collaborate with the CSP to recover real model components using Algorithm \ref{alg:fedpls_recover}.
  }
\caption{P3LS Training}\label{alg:fedpls_training}
\end{algorithm*}

\subsection{Secure data aggregation}
The TA generates two random orthogonal matrices $\bm{A} \in \mathbb{R}^{m \times m}$, $\bm{H} \in \mathbb{R}^{n \times n}$, $\bm{G} \in \mathbb{R}^{l \times l}$ using the block-based efficient mask generation method \cite{chai2022}, and then it splits $\bm{H}^\top$ into $[\bm{H}_{1}^{\top}, ..., \bm{H}_{g}^{\top}]$ where $\bm{H}_i^\top \in \mathbb{R}^{n \times n_{i}}$.

FC-$i$ downloads $\bm{A}$ and $\bm{H}_i$ from the TA, masks their private $\bm{X}$-block such that $\bm{X}_{i}' = \bm{A}\bm{X}_{i}\bm{H}_{i}$, and sends $\bm{X}_{i}'$ to the CSP. At the same time, the LC downloads $\bm{G}$ from the TA and masks the $\bm{Y}$-block through $\bm{Y}' = \bm{A}\bm{Y}\bm{G}$, and sends $\bm{Y}'$ to the CSP.


The CSP aggregates $\bm{X}'$ which is the sum of all $\bm{X}_{i}'$. According to the rule of matrix block multiplication, we have:
\begin{align*}
    \bm{X}' = \sum_{i=1}^{g} \bm{X}_{i}' = \sum_{i=1}^{g} \bm{A}\bm{X}_{i}\bm{H}_{i} &= \bm{A}[\bm{X}_1, ..., \bm{X}_g][\bm{H}_{1}^{\top}, ..., \bm{H}_{g}^{\top}]^\top\\
    &= \bm{A}\bm{X}\bm{H}
\end{align*}


After this step, even though the CSP receives $\bm{X}'$ and $\bm{Y}'$, it has been proved that the CSP cannot recover the original data \cite{chai2022}.

\subsection{Sequential matrix decomposition}

During this step, the latent variables are sequentially estimated. In what follows, we outline the methodology involved in this process and define the relations between the outcomes of P3LS and PLS. As described in \ref{alg:fedpls_training}, in the first iteration, we have
\begin{align*}
    \bm{E}' &:= \bm{X}' = \bm{A}\bm{E}\bm{H} \\
    \bm{F}' &:= \bm{Y}' = \bm{A}\bm{F}\bm{G}
\end{align*}

with the cross-product matrix
\begin{equation*}
    \bm{S}' = \bm{E}'^\top\bm{F}'= \bm{H}^\top\bm{E}^\top\bm{A}^\top\bm{A}\bm{F}\bm{G}.
\end{equation*}

Since $\bm{A}$ is an orthogonal matrix, $\bm{A}^\top\bm{A} = I$, therefore
\begin{equation*}
    \bm{S}' = \bm{H}^\top\bm{E}^\top\bm{F}\bm{G} = \bm{H}^\top\bm{S}\bm{G}.
\end{equation*}

Note that both $\bm{H}^\top$ and $\bm{G}$ are orthogonal matrices. Thus, according to Equation \ref{eq:fedsvd}, the results of performing SVD on $\bm{S}$ and $\bm{S}'$ are convertible.

Next, we calculate X-scores and Y-scores. Since $\bm{H}^\top$ and $\bm{G}$ are orthogonal matrices, $\bm{H}\bm{H}^\top = I$ and $\bm{G}\bm{G}^\top = I$.
\begin{align*}
    \bm{t}' &= \bm{E}'\bm{w}' = \bm{A}\bm{E}\bm{H}\bm{H}^\top\bm{w} = \bm{A}\bm{E}\bm{w} = \bm{A}\bm{t} \\
    \bm{u}' &= \bm{F}'\bm{v}' = \bm{A}\bm{F}\bm{G}\bm{G}^\top\bm{v} = \bm{A}\bm{F}\bm{v} = \bm{A}\bm{u}
\end{align*}

The X-scores $\bm{t}'$ are normalised:
\begin{equation*}
    \bm{t}' = \frac{\bm{t}'}{\sqrt{\bm{t}'^\top\bm{t}'}} = \frac{\bm{A}\bm{t}}{\sqrt{\bm{t}^\top\bm{A}^\top\bm{A}\bm{t}}} = \frac{\bm{A}\bm{t}}{\sqrt{\bm{t}^\top\bm{t}}}.
\end{equation*}

Next, X- and Y-loadings are obtained by regressing $\bm{E}'$ and $\bm{F}'$ against the same vector $\bm{t}'$:
\begin{align*}
    \bm{p}' &= \bm{E}'^\top\bm{t}' =  \bm{H}^\top\bm{E}^\top\bm{A}^\top\bm{A}\bm{t} = \bm{H}^\top\bm{E}^\top\bm{t} = \bm{H}^\top\bm{p}\\
    \bm{q}' &= \bm{F}'^\top\bm{t}' =  \bm{G}^\top\bm{F}^\top\bm{A}^\top\bm{A}\bm{t} = \bm{H}^\top\bm{F}^\top\bm{t} = \bm{G}^\top\bm{q}
\end{align*}

Finally, $\bm{E}'$ and $\bm{F}'$ are deflated:
\begin{align*}
    \bm{E}'_{i+1} = \bm{E}'_{i} - \bm{t}'\bm{p}'^\top = \bm{A}\bm{E}_{i}\bm{H} - \bm{A}\bm{t}\bm{p}^\top\bm{H} &= \bm{A}(\bm{E}_{i} - \bm{t}\bm{p}^\top)\bm{H}\\
    &=  \bm{A}\bm{E}_{i+1}\bm{H}\\
    \bm{F}'_{i+1} = \bm{F}'_{i} - \bm{t}'\bm{q}'^\top = \bm{A}\bm{F}_{i}\bm{G} - \bm{A}\bm{t}\bm{q}^\top\bm{G} &= \bm{A}(\bm{F}_{i} - \bm{t}\bm{p}^\top)\bm{G} \\
    &=  \bm{A}\bm{F}_{i+1}\bm{G}
\end{align*}

The estimation of the next latent variables then can start from the SVD of the cross-product matrix $\bm{S}'_{i+1} = \bm{E}'^\top_{i+1}\bm{F}'_{i+1}= \bm{H}^\top\bm{S}_{i+1}\bm{G}$. After every iteration, the vectors $\bm{w}'$, $\bm{t}'$, $\bm{p}'$ and $\bm{q}'$ are saved as columns to form matrices $\bm{W'}$, $\bm{T}'$, $\bm{P}'$ and $\bm{Q}'$, respectively.

The X-rotation matrix $\bm{R}'$ is defined as:
\begin{align*}
    \bm{R}' &= \bm{W}'(\bm{P}'^\top\bm{W}')^{-1}\\
    & = \bm{H}^\top\bm{W}(\bm{P}^\top\bm{H}\bm{H}^\top\bm{W})^{-1}\\
    & =\bm{H}^\top\bm{W}(\bm{P}^\top\bm{W})^{-1} \\
    & =\bm{H}^\top\bm{R}
\end{align*}

Meanwhile, the coefficient matrix $\bm{B}'$ is defined as:
\begin{equation*}
    \bm{B}' = \bm{R}'\bm{Q}'^\top = \bm{H}^\top\bm{R}\bm{Q}^\top\bm{G} = \bm{H}^\top\bm{B}\bm{G}
\end{equation*}

In summary, the relationships between the PLS and P3LS model components are:

\begin{align} 
    \bm{T} &= \bm{A}^\top\bm{T}' \label{eq:recover_T}\\
    \bm{W} &= \bm{H}\bm{W}'\\
    \bm{P} &= \bm{H}\bm{P}'\\
    \bm{Q} &= \bm{G}\bm{Q}' \label{eq:recover_Q}\\
    \bm{R} &= \bm{H}\bm{R}'\\
    \bm{B} &= \bm{H}\bm{B}'\bm{G}^\top 
\end{align}

\subsection{Decryption}
A challenge that P3LS faces is how to remove the masks securely from the encrypted results computed by the CSP. Motivated by FedSVD, we propose Algorithm \ref{alg:fedpls_recover} to recover the real results of P3LS.

Since all FCs have the keys $\bm{A}$, they can download $\bm{T}'$ and get the real $\bm{T}$ using Equation \ref{eq:recover_T}. Similarly, the LC can get $\bm{Q}'$ from the CSP and recover $\bm{Q}$ using Equation \ref{eq:recover_Q}. On the other hand, since FC-$i$ only possesses a portion of $\bm{H}$, it can recover $\bm{W}_i$, $\bm{P}_i$, and $\bm{B}_i$ through:
\begin{align}
    \bm{W}_i &= \bm{H}_i\bm{W}'\\
    \bm{P}_i &= \bm{H}_i\bm{P}'\\
    \bm{B}_i &= \bm{H}_i\bm{B}'\bm{G}^\top
\end{align}

However, during the computation, we want to guarantee the confidentiality of $\bm{H}_i$, $\bm{G}$, $\bm{W}'$, $\bm{P}'$, $\bm{Q}'$, and $\bm{B}'$, i.e., the FCs should not directly get $\bm{W}'$, $\bm{P}'$, $\bm{Q}'$, and $\bm{B}'$ and the CSP should not be able to learn $\bm{H}_i$ and $\bm{G}$. If the CSP knew $\bm{H}_i$ and $\bm{G}$, they could easily estimate $\bm{W}_i$, $\bm{P}_i$, $\bm{Q}$, and $\bm{B}_i$.

We propose to first mask $\bm{H}_i \in \mathbb{R}^{n_i \times n}$ using a locally-generated random matrix $\bm{C}_i \in \mathbb{R}^{n_i \times n_i}$ according to Equation \ref{eq:mask_Hi}. Next, FC-$i$ sends $[\bm{H}_i]^{\bm{C}_i}$ (i.e., the masked matrix) to the CSP, which will subsequently compute $[\bm{W}_i]^{\bm{C}_i}$ according to Equation \ref{eq:compute_Wi} and send $[\bm{W}_i]^{\bm{C}_i}$ back to FC-$i$. Then FC-$i$ can remove the random mask according to Equation \ref{eq:remove_mask_Hi} and get the final result (i.e., $\bm{W}_i$). 
\begin{align}
    [\bm{H}_i]^{C} &= \bm{C}_i\bm{H}_i \label{eq:mask_Hi} \\
    [\bm{W}_i]^{C} &= [\bm{H}_i]^{\bm{C}_i}\bm{W}' (= \bm{C}_i\bm{W}_i) \label{eq:compute_Wi}\\
    \bm{W}_i &= \bm{C}_i^{-1}[\bm{W}_i]^{\bm{C}_i} \label{eq:remove_mask_Hi}
\end{align}
This same approach can be applied to recover $\bm{P}_i$.

It is more complicated when it comes to recovering $\bm{B}_i$ since $\bm{B}_i$ is protected by two keys, one of which is known only to the LC (.i.e, $\bm{G}$). Our idea is to replace $\bm{G}$ with a common key known to all FCs, then perform a similar procedure as recovering $\bm{W}_i$. First, the TA generates a random matrix $\bm{N} \in \mathbb{R}^{n \times n}$. Next, the LC downloads $\bm{N}$ and masks $\bm{G}^\top$ through $[\bm{G}^\top]^{\bm{N}} = \bm{G}^\top\bm{N}$, and then sends $[\bm{G}^\top]^{\bm{N}}$ to the CSP. Then, the CSP calculates $[\bm{B}_{i}]^{\bm{C}_{i}}$ by the operation $[\bm{B}_{i}]^{\bm{C}_{i}} = [\bm{H}_{i}]^{\bm{C}_{i}}\bm{B}'[\bm{Q}^\top]^{\bm{N}}$. Mathematically, $[\bm{H}_{i}]^{\bm{C}_{i}}\bm{B}'[\bm{Q}^\top]^{\bm{N}} = \bm{C}_i\bm{H}_i\bm{B}'\bm{G}^\top\bm{N} = \bm{C}_{i}\bm{B}_{i}\bm{N}$. After that, the CSP sends $[\bm{B}_{i}]^{\bm{C}_{i}}$ to FC-$i$. Finally, FC-$i$ downloads $\bm{N}$ from the TA and recovers $\bm{B}_{i}$ through $\bm{B}_{i} = \bm{C}_{i}^{-1}[\bm{B}_{i}]^{\bm{C}_{i}}\bm{N}^{-1}$.

  
\begin{algorithm*}[hbt!]
\DontPrintSemicolon
  
  \Fn{\PPPLSRecover{\textbf{\empty}}}{
        \KwTA(do:){
            Generate a random matrix $\bm{N} \in \mathbb{R}^{l \times l}$
         }
        \KwDH(do:){
            \For{$i = 1 \rightarrow g$, FC-$i$}{
                Generate a random matrix $\bm{C}_{i} \in \mathbb{R}^{n_{i} \times n_{i}}$\\
                Mask $\bm{H}_i$ through: $[\bm{H}_{i}]^{\bm{C}_{i}} = \bm{C}_{i}\bm{H}_{i}$\\
                Send $[\bm{H}_{i}]^{\bm{C}_{i}}$ to the CSP
            }
        }
        \KwLC(do:){
            Download $\bm{N}$ from the TA\\
            Mask $\bm{G}^\top$ through: $[\bm{G}^\top]^{\bm{N}} = \bm{G}^\top\bm{N}$\\
            Send $[\bm{G}^\top]^{\bm{N}}$ to the CSP
         }
        \KwCSP(wait to receive data then do:){
            \If{Receive $[\bm{H}_{i}]^{C}$}
            {
                Compute the following:
                \begin{equation*}
                    [\bm{W}_{i}]^{\bm{C}_{i}} = [\bm{H}_{i}]^{\bm{C}_{i}}\bm{W}', \quad
                    [\bm{P}_{i}]^{\bm{C}_{i}} = [\bm{H}_{i}]^{\bm{C}_{i}}\bm{P}', \quad
                    [\bm{B}_{i}]^{\bm{C}_{i}} = [\bm{H}_{i}]^{\bm{C}_{i}}\bm{B}'[\bm{Q}^\top]^{\bm{N}}
                \end{equation*}
                
                Send $[\bm{W}_{i}]^{\bm{C}_{i}}$, $[\bm{P}_{i}]^{\bm{C}_{i}}$, and $[\bm{B}_{i}]^{\bm{C}_{i}}$ to FC-$i$  
            }
            Send $\bm{Q}'$ only to the LC
         }
         \KwLC(do:){
            Recover $\bm{Q}$ by: $\bm{Q} = \bm{G}\bm{Q}'$
         }
         \KwDH(do:){
           \For{$i = 1 \rightarrow g$, FC-$i$}{
                Download $\bm{N}$ from the TA\\
                Get $[\bm{W}_{i}]^{C}$, $[\bm{P}_{i}]^{C}$, and $[\bm{B}_{i}]^{C}$ from the CSP \\
                Recover $\bm{W}_{i}$, $\bm{P}_{i}$, and $\bm{B}_{i}$ through: 
                \begin{equation*}
                    \bm{W}_{i} = \bm{C}_{i}^{-1}[\bm{W}_{i}]^{\bm{C}_{i}}, \quad
                    \bm{P}_{i} = \bm{C}_{i}^{-1}[\bm{P}_{i}]^{\bm{C}_{i}}, \quad
                    \bm{B}_{i} = \bm{C}_{i}^{-1}[\bm{B}_{i}]^{\bm{C}_{i}}\bm{N}^{-1}
                \end{equation*}
           }
         }
  }
\caption{P3LS Recover}\label{alg:fedpls_recover}
\end{algorithm*}

\subsection{Incentive mechamisms}

Incentive mechanisms are pivotal in real-world applications, to promote participant engagement, enhance collaboration, and ensure the truthful sharing of data. Intuitively, data providers with high contributions deserve a better payoff. Considering only the volume of contributed data is certainly not sufficient because data quality is equally essential as data quantity in building robust and effective models. 



The explained variance, represented as $R^2_i$, quantifies the amount of variance present in the data block $\bm{X}_i$ that can be described by the latent variables $\bm{T}$ within the P3LS model. This calculation is derived from the $\bm{X}$-loadings matrix $\bm{P}_i$ and can be expressed through the following equation:


\begin{equation}
    R^2_{X_i} = \frac{SS(\bm{P}_i)}{SS(X)} = \frac{SS(\bm{P}_i)}{(m-1)n}
\end{equation}

Similarly, the explained variance in $\bm{Y}$ can be quantified by the following equation:


\begin{equation}
    R^2_Y = \frac{SS(\bm{Q})}{SS(\bm{Y})} = \frac{SS(\bm{Q})}{(m-1)l}
\end{equation}


where $SS(\cdot)$ denotes the sum of squares of the term in parentheses \cite{abdi2003}. Since $m$ and $l$ are known to all data holders, the above equations can be calculated locally. 

To evaluate the relevance of data, it is intuitive to examine how well each data block (i.e., $\bm{X}_i$) is able to predict the target block (i.e., $\bm{Y}$). The standard way to assess predictive power involves determining how much variance in $\bm{Y}$ can be accounted for by $\bm{X}_i$. Algorithm \ref{alg:fedpls_ev} outlines the steps to compute this metric.

Mathematically, $R^2_{X_iY}$ can be computed through the following equation:
\begin{equation}\label{eq:R2_xy}
    R^2_{X_iY} = 1 - \frac{SS(\bm{Y} - \hat{\bm{Y}}_i)}{SS(\bm{Y})} = 1 - \frac{SS(\bm{Y} - \bm{X}_i\bm{B}_i)}{ml}
\end{equation}

To keep $\hat{\bm{Y}}_i$ and $\bm{Y}$ private to FC-$i$ and the LC, respectively, we propose a procedure that involves the TA and the CSP. First, the TA generates two orthogonal matrices $\bm{M} \in \mathbb{R}^{m \times m}$ and $\bm{N} \in \mathbb{R}^{l \times l}$. Next, all FCs download $\bm{M}$ and $\bm{N}$ from TA and use them to mask their local predictions $\hat{\bm{Y}}_i$ via $\hat{\bm{Y}}'_i = \bm{M}\hat{\bm{Y}}_i\bm{N}$ and send $\hat{\bm{Y}}'_i$ to the CSP. Meanwhile, the LC mask $\bm{Y}$ via $\bm{Y}' = \bm{M}\bm{Y}\bm{N}$ and sends $\bm{Y}'$ to the CSP. Then, the CSP calculates the residual matrix $\bm{E}'_i = \bm{Y}' - \hat{\bm{Y}}'_i$ which is equivalent of $\bm{M}(\bm{Y} - \hat{\bm{Y}}_i)\bm{N}$ or $\bm{M}\bm{E}_i\bm{N}$. After that, the CSP performs SVD on $\bm{E}'^\top_i\bm{E}'_i$ ($=\bm{N}^\top\bm{E}^\top_i\bm{E}_i\bm{N}$). Since $\bm{N}^\top$ and $\bm{N}$ are two random orthogonal matrices, it has been proved that the SVD results of $\bm{N}^\top\bm{E}_i^\top\bm{E}_i\bm{N}$ and $\bm{E}_i^\top\bm{E}_i$ share the same eigenvalues \cite{chai2022}. Thus, the CSP can calculate the sum of eigenvalues of $\bm{E}_i^\top\bm{E}_i$, which is equivalent to the trace of $\bm{E}_i^\top\bm{E}_i$ and $SS(\bm{E}_i)$. Finally, the CSP forwards $SS(\bm{E}_i)$ to FC-$i$, who proceeds to calculate $R^2_{X_iY}$ using Equation \ref{eq:R2_xy}.

\begin{algorithm}[hbt!]
\DontPrintSemicolon
  
  \Fn{\PPPLSRecover{\textbf{\empty}}}{
        \KwTA(do:){
            Generate two orthogonal matrices $\bm{M} \in \mathbb{R}^{m \times m}$ and $\bm{N} \in \mathbb{R}^{l \times l}$
         }
        \KwDH(do:){
            \For{$i = 1 \rightarrow g$, FC-$i$}{
                Download $\bm{M}$ and $\bm{N}$ from the TA\\
                Calculate $\hat{\bm{Y}}_i$ through: $\hat{\bm{Y}}_i = \bm{X}_i\bm{B}_i$\\
                Mask $\hat{\bm{Y}}_i$ through: $\hat{\bm{Y}}'_i = \bm{M}\hat{\bm{Y}}_i\bm{N}$\\
                Send $\hat{\bm{Y}}'_i$ to the CSP
            }
        }
        \KwLC(do:){
            Download $\bm{M}$ and $\bm{N}$ from the TA\\
            Mask $\bm{Y}$ through: $\bm{Y}' = \bm{M}\bm{Y}\bm{N}$\\
            Send $\bm{Y}'$ to the CSP
         }
        \KwCSP(wait to receive data then do:){
            Compute the residual matrix: $\bm{E}'_i = \bm{Y}' - \hat{\bm{Y}}'_i$\\
            Perform SVD on $\bm{E}'^\top_i\bm{E}'_i$ and extract all eigenvalues ($\lambda_i$)\\
            Calculate $SS(\bm{E}'_i)$ by: $SS(\bm{E}'_i) = \sum \lambda_i$\\
            Send $SS(\bm{E}'_i)$ to FC-$i$
         }
         \KwDH(do:){
            \For{$i = 1 \rightarrow g$, FC-$i$}{
                Download $SS(\bm{E}'_i)$ from the CSP\\
                Calculate $R^2_{X_iY}$ by: $R^2_{X_iY} = 1 - SS(\bm{E}'_i)/ml$
            }
        }
  }
\caption{Variance explained in $\bm{Y}$ by $\bm{X}_i$}\label{alg:fedpls_ev}
\end{algorithm}


\section{Security analysis} \label{sec:security_analysis}

As P3LS incorporates FedSVD at its core, it inherits the security assurances provided by FedSVD's proof of security \cite{chai2022}, which includes the following:

\begin{itemize}
    \item The TA only knows the keys and never sees the real/encrypted data.
    \item The CSP solely receives encrypted data and learns encrypted model components within the framework. Importantly, it has been established that it is not feasible to directly deduce or infer the raw data from the encrypted data within this context.
    \item The data holders only see their private data and the relevant components of the model that correspond to the data that they contribute.
\end{itemize}

Despite their potential usefulness, specific model components could be withheld from the data holders to ensure the preservation of security and confidentiality within the framework.

The decision not to disclose the local rotation matrix $\bm{R}_i$ to FC-$i$ is intentional. By withholding $\bm{R}_i$, it prevents FC-$i$ from accessing both $\bm{B}_i$ and $\bm{R}_i$ simultaneously, as knowledge of both enables estimation of $\bm{Q}^\top$ through the formula $\bm{Q}^\top = (\bm{R}_i^\top\bm{R}_i)^{-1}\bm{R}_i^\top\bm{B}_i$. As $\bm{Q}$ contains the loadings of the response variables, it is crucial for this information to be known exclusively by the LC. While hiding $\bm{R}_i$, we allow FC-$i$ to learn $\bm{B}_i$ due to specific functional purposes within the framework:

\begin{itemize}
    \item Joint estimation of target variables: FC-$i$ can still jointly estimate the target variables.
    \item Insights into cross-organizational process interactions: Access to $\bm{B}_i$ allows FC-$i$ to learn how their process variables interact with the product quality of the downstream company. Such information might be useful when they try to optimize their processes and want to take the downstream company's KPIs into account.
\end{itemize}

In addition, we strategically decide to hide the local scores $\bm{T}_i$ from data holders to prevent potential inference of the local rotation matrix ($\bm{R}_i$). 

\begin{equation*}
    \bm{T}_i = \bm{X}_i\bm{R}_i \rightarrow \bm{R}_i=(\bm{X}_i^\top \bm{X}_i)^{-1} \bm{X}_i^\top\bm{T}_i
\end{equation*}

Even if the left inverse isn't unique, it doesn't necessarily imply that data holders are unable to find a suitable solution. As highlighted earlier, the primary reason for restricting access to $\bm{R}_i$ from FCs is due to their potential ability to estimate $\bm{Q}$, which encompasses crucial information related to the target variables.


\section{Applications of P3LS} \label{sec:applications}

As a variant of PLS, P3LS naturally lends itself to applications where PLS is commonly employed (e.g., in MSPC). One notable advantage of utilizing P3LS lies in its ability to account for variations from preceding stages. Consequently, it has the potential to enhance the quality control performance \cite{shi2009}. Some of the applications that we consider relevant for P3LS include:

\textbf{Process monitoring:} PLS is widely acknowledged as one of the most popular methods in process monitoring \cite{ge2012}. P3LS can be directly employed for this purpose. Algorithm \ref{alg:fedpls_inference} outlines the steps involved in estimating the global X-scores matrix ($\bm{T}$). By utilizing $\bm{T}$ along with the residual matrix ($\bm{\Theta}_i$), it becomes feasible to compute commonly used monitoring metrics such as Hotelling's $T^2$ and $Q$-statistics, and the contributions from original variables. These indices serve as valuable tools in process monitoring and fault detection and diagnosis \cite{macgregor1995}.

\textbf{Soft sensors}: The purpose of an inferential sensor is to deduce a property that is challenging or expensive to measure directly — usually a laboratory measurement or a costly parameter — by leveraging a blend of process data and software-driven algorithms. Once validated, a soft sensor can significantly minimize process costs by enabling swift feedback control over the estimated property. This control mechanism aids in reducing the production of off-specification products. Moreover, an additional benefit of soft sensors is their capacity to curtail the need for extensive laboratory sampling, thus reducing manpower costs. Soft sensors leveraging latent variables predominantly rely on PLS models. Algorithm \ref{alg:fedpls_inference} outlines the steps involved in generating predictions for unseen data. However, unlike the global score matrix $\bm{T}$, the prediction of the response variables is constrained to the Label Contributors (LC) exclusively.

\textbf{Improved process understanding}: The regression coefficients allow all data holders to grasp the correlation of their process variables with some particular KPIs of the final product. This acquired knowledge proves immensely valuable, particularly during process optimization, as it provides insights into how their variables influence the desired outcomes. Moreover, this valuable understanding is only accessible to the participants through their involvement in the data federation.

\begin{algorithm*}[hbt!]
\DontPrintSemicolon
  
  \KwInput{$\bm{X} = [\bm{X}_1, ..., \bm{X}_g] \in \mathbb{R}^{m \times n}$, where $\bm{X}_i \in \mathbb{R}^{m \times n_i}$}
  \KwOutput{$\bm{T}$, $\hat{\bm{Y}}$}
  \Fn{\PPPLSPredict{$[\bm{X}_1, ..., \bm{X}_g]$}}{
         \KwTA(do:){
            Generate a random matrix $\bm{M} \in \mathbb{R}^{m \times m}$
         }
         \KwDH(do:){
           \For{$i = 1 \rightarrow g$, FC-$i$}{
                Download $\bm{M}$ from the TA \\
                Make local predictions: $\hat{\bm{Y}}_i = \bm{X}_i\bm{B}_{i}$\\
                Mask local predictions: $\hat{\bm{Y}}'_i = \bm{M}\hat{\bm{Y}}_i$\\
                Encrypt local data using $\bm{M}$ and $\bm{H}_i$: $\bm{X}'_i = \bm{M}\bm{X}_i\bm{H}_i$\\
                Send $\hat{\bm{Y}}'_i$ and $\bm{X}'_i$ to the CSP
           }
         }
        \KwCSP(do:){
            Aggregate $\bm{X}'$: $\bm{X}' = \sum_{i=1}^{g} \bm{X}'_i \quad (= \bm{M}\bm{X}\bm{H})$\\
            Aggregate $\hat{\bm{Y}}' = \sum_{i=1}^{g} \hat{\bm{Y}}'_{i} \quad (=\bm{M}\hat{\bm{Y}})$\\
            Calculate the masked global scores: $\bm{T}' = \bm{X}'\bm{R}' \quad (=\bm{M}\bm{X}\bm{H}\bm{H}^\top\bm{R}=\bm{M}\bm{X}\bm{R}=\bm{M}\bm{T})$
         }
         \KwDH(do:){
           \For{$i = 1 \rightarrow g$, FC-$i$}{
                Download $\bm{T}'$ from the CSP\\
                Recover the real global scores: $\bm{T} = \bm{M}^{-1}\bm{T}'$
           }
         }
         \KwLC(do:){
            Download $\hat{\bm{Y}}'$ from the CSP\\
            Recover the real predictions: $\hat{\bm{Y}} = \bm{M}^{-1}\hat{\bm{Y}}'$
         }
  }
\caption{P3LS Inference}\label{alg:fedpls_inference}
\end{algorithm*}

\section{Experimental evaluation} \label{sec:experiment}

To assess the efficacy of P3LS, an experiment utilizing synthetic datasets has been designed. The primary objective of this experiment is to address the following questions:
\begin{itemize}
    \item Q1: What is the performance gap between P3LS and a centralized PLS model that lacks privacy protection?
    \item Q2: Does introducing more data from other participants improve the model performance?
\end{itemize}

In essence, Q1 delves into the inherent trade-offs associated with safeguarding privacy within a vertical federated setting, while Q2 examines the potential enhancements resulting from collaborative training involving diverse data sources. We continue by describing the datasets and then the experiment settings. Finally, the experiment results will be employed to address the previously posed questions. 

\subsection{Datasets}

Due to the absence of realistic datasets in VFL as indicated by \cite{liu2022}, we implemented a multistage process simulator outlined in \cite{jin1999} to generate five synthetic datasets of varying sizes. For comprehensive details regarding the simulator itself and the specific configurations adopted for generating these synthetic datasets, an extensive explanation is available in \ref{appendix:mps}. 

\begin{figure}[th]
\includegraphics[scale=0.4]{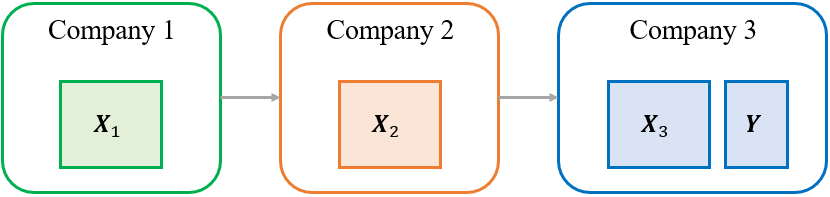}
\centering
\caption{The structure of the simulated datasets.}
  \label{fig:simulated_dataset}
\end{figure}

Each of the generated datasets simulates a three-stage process. In the context of a vertical federated scenario, the assumption made is that there are three distinct manufacturing companies, with each company controlling and possessing data from a specific stage of the process. Although quality characteristics are recorded across all stages, the primary objective of the data federation is to construct a predictive model solely for the output quality of the final stage. The process variables specific to company $i$'s are denoted as $\bm{X}_i$, while the quality variables of the final product are represented as $\bm{Y}$. An illustration of the structure of the simulated dataset is provided in Figure \ref{fig:simulated_dataset}. Furthermore, Table \ref{table:synthetic_data} presents a summary detailing the dataset owned by each individual company.

\begin{table}[ht]
\centering
\caption{Process variables and company assignment.}
\label{table:synthetic_data}
\begin{tabular}{|c|l|l|}
\hline
Dataset & \multicolumn{1}{c|}{$\bm{X}$-blocks} & \multicolumn{1}{c|}{$\bm{Y}$-block} \\ \hline
&&\\[-1em]
\#1 & \begin{tabular}[c]{@{}l@{}}Comp 1: $\bm{X}_1 \in \mathbb{R}^{1000 \times 10}$\\ Comp 2: $\bm{X}_2 \in \mathbb{R}^{1000 \times 20}$\\ Comp 3: $\bm{X}_3 \in \mathbb{R}^{1000 \times 20}$\end{tabular} & \multirow{5}{*}{\begin{tabular}[c]{@{}l@{}}Comp 3:\\ $\bm{Y} \in \mathbb{R}^{1000 \times 7}$\end{tabular}} \\ \cline{1-2}&&\\[-1em]
\#2 & \begin{tabular}[c]{@{}l@{}}Comp 1: $\bm{X}_1 \in \mathbb{R}^{1000 \times 20}$\\ Comp 2: $\bm{X}_2 \in \mathbb{R}^{1000 \times 40}$\\ Comp 3: $\bm{X}_3 \in \mathbb{R}^{1000 \times 40}$\end{tabular} &  \\ \cline{1-2}&&\\[-1em]
\#3 & \begin{tabular}[c]{@{}l@{}}Comp 1: $\bm{X}_1 \in \mathbb{R}^{1000 \times 50}$\\ Comp 2: $\bm{X}_2 \in \mathbb{R}^{1000 \times 100}$\\ Comp 3: $\bm{X}_3 \in \mathbb{R}^{1000 \times 100}$\end{tabular} &  \\ \cline{1-2}&&\\[-1em]
\#4 & \begin{tabular}[c]{@{}l@{}}Comp 1: $\bm{X}_1 \in \mathbb{R}^{1000 \times 100}$\\ Comp 2: $\bm{X}_2 \in \mathbb{R}^{1000 \times 200}$\\ Comp 3: $\bm{X}_3 \in \mathbb{R}^{1000 \times 200}$\end{tabular} &  \\ \cline{1-2}&&\\[-1em]
\#5 & \multicolumn{1}{c|}{\begin{tabular}[c]{@{}c@{}}Comp 1: $\bm{X}_1 \in \mathbb{R}^{1000 \times 200}$\\ Comp 2: $\bm{X}_2 \in \mathbb{R}^{1000 \times 400}$\\ Comp 3: $\bm{X}_3 \in \mathbb{R}^{1000 \times 400}$\end{tabular}} &  \\ \hline
\end{tabular}
\end{table}

\subsection{Experiment settings}\label{analysis}

All runs were conducted on a workstation using an 11th Gen Intel(R) Core(TM) i7-1185G7 with four cores at 3.00GHz, 32.0 GB RAM, and PyCharm 2022.1.

Each private dataset was randomly partitioned into three separate subsets: training set (60\%), validation set (20\%), and test set (20\%). The training set was dedicated to constructing the models. The validation set was utilized for optimizing the number of latent variables in the constructed models. In particular, the optimal number is the one that returns the highest accuracy on the validation set. The test set was designated to estimate the model's performance on unseen data.

For each of the generated datasets, we constructed three distinct models, each reflecting a scenario commonly observed in cross-organizational manufacturing processes:
\begin{itemize}
    \item \textbf{CenPLS}:  A PLS model trained by the last company using the centralized dataset. It simulates a scenario where a single company has access to and utilizes all available data along the value chain. 
    
    \item \textbf{LocalPLS}: A PLS model trained exclusively on the private dataset of the last company (i.e., $\bm{X}_3^{train}$ and $\bm{Y}^{train}$).
    
    \item \textbf{P3LS}: A P3LS model trained on all available data along the value chain without direct data sharing between companies.
\end{itemize}

A detailed description of the data used for developing and testing each model is described in Table \ref{table:general_settings_data_split}. To answer Q1 and Q2, we considered the following factors:
\begin{itemize}
    \item Losslessness: Even though it has been proved that the outcomes of CenPLS and P3LS are equivalent, we wanted to empirically compare the components of CenPLS and P3LS models using the mean squared distance.
    \item Prediction performance: We measured the $R^2$ of P3LS, CenPLS, and LocalPLS when applying these models to the test set.
    \item Computation time: We measured the run time of all three models when making inferences. 
\end{itemize}

Since P3LS involves using random matrices to gain a comprehensive evaluation, we repeated the experiment 100 times for each dataset. 

\begin{table*}[ht!]
\centering
\caption{Training set, validation set, and test set for each model.}
\begin{tabular}{| >{\centering}p{1.5cm} | >{\centering}p{4cm} | >{\centering}p{3.5cm} | >{\centering\arraybackslash}p{3.5cm} |} 
 \hline
 Model & Training Data & Validation Data & Test Data \\ [0.5ex]
 \hline
 CenPLS & $\bm{X}^{train}$, $\bm{Y}^{train}$ & $\bm{X}^{val}$, $\bm{Y}^{val}$ & $\bm{X}^{test}$, $\bm{Y}^{test}$ \\ 
 \hline
 LocalPLS & $\bm{X}_{3}^{train}$, $\bm{Y}^{train}$ & $\bm{X}_{3}^{val}$, $\bm{Y}^{val}$ & $\bm{X}_{3}^{test}$, $\bm{Y}^{test}$ \\ 
 \hline
 P3LS & $\bm{X}_{1}^{train}$, $\bm{X}_{2}^{train}$, $\bm{X}_{3}^{train}$, $\bm{Y}^{train}$ & $\bm{X}_{1}^{val}$, $\bm{X}_{2}^{val}$, $\bm{X}_{3}^{val}$, $\bm{Y}^{val}$ & $\bm{X}_{1}^{test}$, $\bm{X}_{2}^{test}$, $\bm{X}_{3}^{test}$, $\bm{Y}^{test}$ \\ 
 \hline
\end{tabular}
\label{table:general_settings_data_split}
\end{table*}

\subsection{Results}

\begin{figure*}[th]
\includegraphics[scale=0.32]{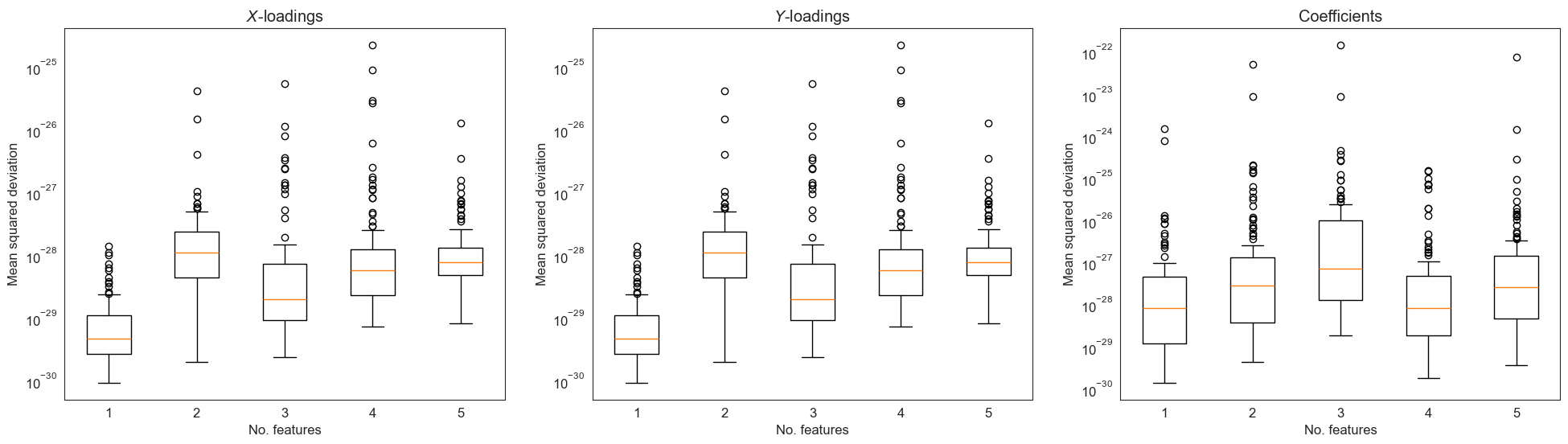}
\centering
\caption{Compare model components of CenPLS and P3LS.}
  \label{fig:cenpls_p3ls_components}
\end{figure*}

\begin{figure}[th]
\includegraphics[scale=0.35]{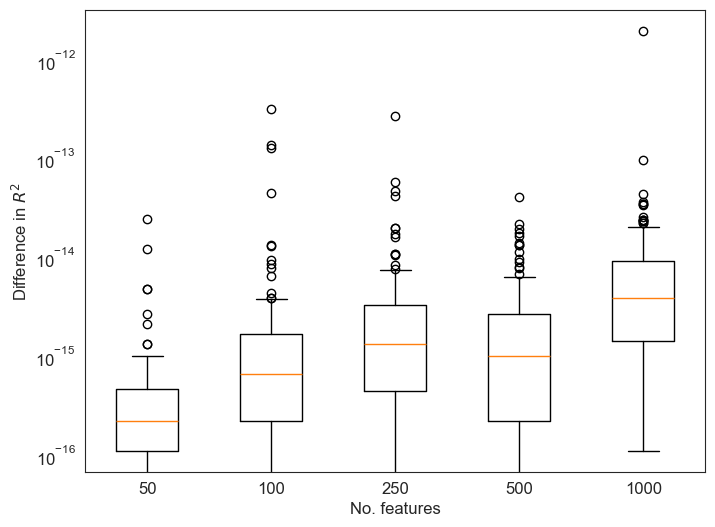}
\centering
\caption{Difference in test $R^2$ between CenPLS and P3LS.}
  \label{fig:cenpls_p3ls_diff_r2}
\end{figure}

The similarity between P3LS and CenPLS is empirically confirmed when comparing corresponding model components. Figure \ref{fig:cenpls_p3ls_components} displays the mean squared distance between different components of the two models across all datasets. It illustrates that the difference between the two models is negligible. Consequently, as shown in Figure \ref{fig:cenpls_p3ls_diff_r2}, the difference in their $R^2$ scores is insignificant.

It can also be seen in Figure \ref{fig:cenpls_p3ls_diff_r2} that for the same dataset, even though the results of different runs are not the same, the variations are ignorable. It is important to note that P3LS is theoretically lossless. However, minor deviations in the experiment might occur due to floating-point number representation in computers. 

\begin{figure}[th]
\includegraphics[scale=0.35]{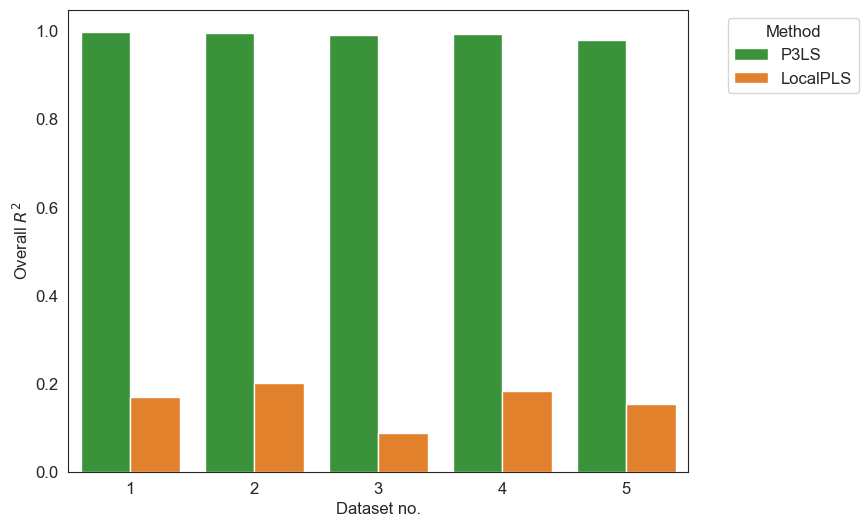}
\centering
\caption{Comparison of performance on the test set of CenPLS, LocalPLS, and P3LS.}
  \label{fig:localpls_p3ls_r2}
\end{figure}

Figure \ref{fig:localpls_p3ls_r2} compares the overall $R^2$ of LocalPLS and P3LS when evaluating these models on the test set. Across all datasets, it shows that P3LS models outperform the local PLS models. It can be explained by the fact that the underlying process models involve variables from all stages. Since LocalPLS models only learn from the last company's data, they miss crucial information needed to make good predictions. However, in practice, there might be cases where the target variables are totally independent of preceding production steps, or the information provided by other data providers is not relevant. In such cases, the advantages of using P3LS might be less significant.

\begin{figure}[th]
\includegraphics[scale=0.35]{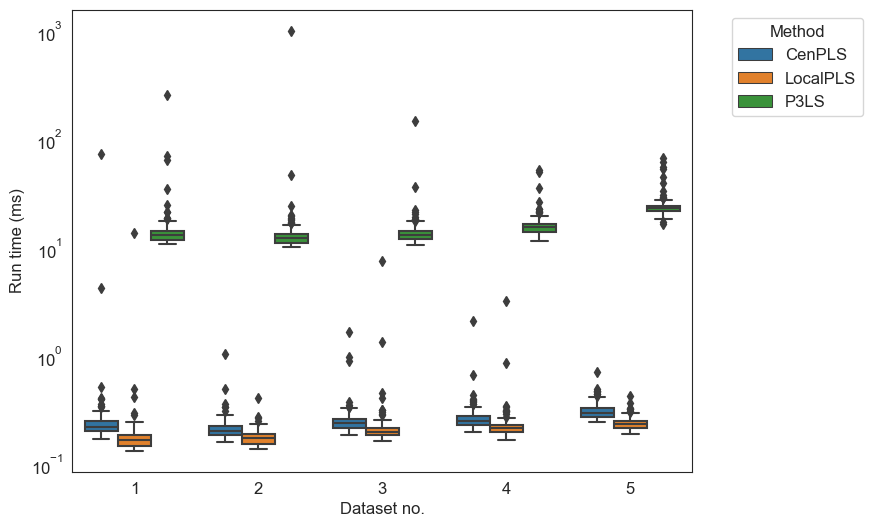}
\centering
\caption{Computation time in the inference phase.}
  \label{fig:run_time}
\end{figure}

Figure \ref{fig:run_time} illustrates the computation time of the three models. As expected, compared to CenPLS or LocalPLS, it took P3LS more time to produce inferences. This results from processing more data and the additional complexity introduced by the data encryption and decryption steps. It should be highlighted that the experiment was performed in simulation mode, and the communication time was not considered. Therefore, the results might differ when deploying in real systems with more advanced hardware.



\section{Discussion and conclusions} \label{sec:discussion}

This paper proposes P3LS, which allows multiple companies to collaboratively build and exploit an integrated cross-organizational process model without revealing private datasets. The empirical experiment on simulated data shows that the proposed approach can yield a model with comparable performance to the conventional approach of centralizing data while protecting the data privacy of the participants. Besides, we also proposed an approach for quantifying the contribution of participants. This approach creates a basis for a profit-sharing scheme and, at the same time, motivates companies to join the federation and contribute high-quality data.

Even though the paper focuses on the vertical setting, the proposed method can also be extended to apply to the horizontal setting. In the horizontal setting, each data holder acts as both a feature contributor and a label contributor as they own both process variables (i.e., $\bm{X}_i$) and response variables (i.e., $\bm{Y}_i$). Under this setting, the shared model components are the weights, loadings, and coefficients. All steps in Algorithm \ref{alg:fedpls_training} are kept the same, except the following:

\begin{itemize}
    \item Step 2-4: Instead of splitting $\bm{H}$, the TA splits $\bm{A}^\top$ into $[\bm{A}_1^\top,...,\bm{A}_g^\top]$ where $\bm{A}_i^\top \in \mathbb{R}^{m_i \times n}$.
    \item Step 6-15: FC $i$ downloads $\bm{A}_i^\top$ and $\bm{H}$ from the TA and computes $\bm{X}'_i = \bm{A}_i\bm{X}_i\bm{H}$, $\bm{Y}'_i = \bm{A}_i\bm{Y}_i\bm{G}$.
    \item Step 17: In addition to aggregating $\bm{X}'$, the CSP needs to aggregates $\bm{Y}' = \sum_{i=1}^{g} \bm{Y}_{i}'$.
\end{itemize}

At the end of the training phase, all data holders can download the encrypted loadings, encrypted weights, and encrypted coefficients from the CSP. Then, they can use the given keys (i.e., $\bm{A}_i$, $\bm{H}$ and $\bm{G}$) to recover the real model components. After that, they can use these model components to estimate scores and predict the response variables of unseen data.

We will also investigate the scenarios where there are multiple label contributors. In particular, multiple companies along the value chain can contribute their response variable blocks (e.g., $\bm{Y}_i$). Ultimately, companies can learn the interaction between their process variables and all other companies' KPIs, not just the last company. This extension will require a change in key generation and encrypted data aggregation and will be in the scope of our future works.

Another extension of P3LS that will be investigated is Multiway P3LS, which can handle batch process data where $\bm{X}_i$ is in a 3D format with the additional dimension of time. In an offline mode, batch data can be unfolded into a 2D format using the batch-wise unfolding \cite{nomikos1995}, and P3LS can be naturally applied. However, it is more challenging to estimate the response variables in online mode. It is because the $\bm{X}_{new}$ matrix is not complete until the end of the batch operation. Suppose it takes $K$ time units to complete a batch. At time interval $k$, the matrix $\bm{X}_{new}$ has only its first $k$ rows complete, and it is missing all the future observations ($K - k$ rows). A solution to this problem is to leverage the ability of PLS to handle missing data. Multiway PLS does this by projecting the already known observations up to time interval $k$ into the reduced space defined by the $\bm{W}$ and $\bm{P}$ matrices \cite{nomikos1995}. A similar approach will be adopted to enable P3LS to operate in online mode.

Furthermore, in our future research and development, we aim to validate the effectiveness of P3LS in real-life datasets, which tend to be more intricate. Factors that are crucial for practical application in real production scenarios, such as infrastructure requirements and computational time, will also go through a comprehensive assessment.

\section*{Acknowledgment}

The research reported in this paper has been partly funded by the Federal Ministry for Climate Action, Environment, Energy, Mobility, Innovation and Technology (BMK), the Federal Ministry for Digital and Economic Affairs (BMDW), and the State of Upper Austria in the frame of SCCH, a center in the COMET - Competence Centers for Excellent Technologies Program managed by the Austrian Research Promotion Agency FFG and the FFG project circPlast-mr (Grant No. 889843). 

\appendix

\section{Multistage process simulator} \label{appendix:mps}

\begin{figure}[th]
\includegraphics[scale=0.45]{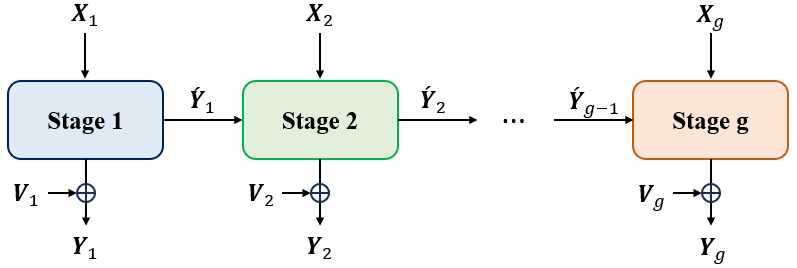}
\centering
\caption{Multistage process simulator diagram.}
  \label{fig:multistage_process_simulator}
\end{figure}

We implemented the multistage process simulator proposed in \cite{shi2009} to generate synthetic data. Figure \ref{fig:multistage_process_simulator} illustrates a typical diagram of a multistage system. At stage-$i$, the process variables and the measured responses are denoted as $\bm{X}_i$ and $\bm{Y}_i$, respectively. Meanwhile, $\bm{V}_i$ represents the measurement noises associated with responses. To introduce the interaction between stages, some responses of upstream stages, denoted as $\bm{\acute{Y}}_{i-1}$, will be used as inputs of stage-$i$'s process model. In this simulator, all stages' process models are set to quadratic functions that can be expressed as:

\begin{equation}
    \bm{Y}_i = \bm{A}_i\bm{U}_{i.lin} + \bm{B}_i\bm{U}_{i.qd} + \bm{V}_i
\end{equation}

where $\bm{U}_{i.lin}$ and $\bm{U}_{i.qd}$ are matrices containing the linear terms and the quadratic terms of the model, respectively. Meanwhile, $\bm{A}_i$ and $\bm{B}_i$ represent the linear and quadratic model coefficients, respectively. The elements of $\bm{A}_i$ and $\bm{B}_i$ are drawn randomly from a specific distribution. While $\bm{A}_i$ is a dense matrix, $\bm{B}_i$ is intentionally made sparse to imitate actual physical process models, often containing only a sparse subset of quadratic terms.

Following are the configurations used for generating Dataset-1:
\begin{itemize}
    \item Stage 1: $\bm{X}_1 \in \mathbb{R}^{1000 \times 10}$, $\bm{Y}_1 \in \mathbb{R}^{1000 \times 5}$, $\bm{A}_1 \sim \mathcal{U}(-1,\,2)$, $\bm{B}_1 \sim \mathcal{U}(-0.01,\,0.02)$. $\bm{A}_1$ has sparsity probability of 0.15. 

    \item Stage 2: $\bm{X}_2 \in \mathbb{R}^{1000 \times 20}$, $\bm{Y}_2 \in \mathbb{R}^{1000 \times 6}$, $\bm{\acute{Y}}_1 \in \mathbb{R}^{1000 \times 3}$, $\bm{A}_2 \sim \mathcal{U}(-3,\,3)$, $\bm{B}_2 \sim \mathcal{U}(-0.03,\,0.03)$. $\bm{A}_2$ has sparsity probability of 0.2. 

    \item Stage 3: $\bm{X}_3 \in \mathbb{R}^{1000 \times 20}$, $\bm{Y}_3 \in \mathbb{R}^{1000 \times 7}$, $\bm{\acute{Y}}_2 \in \mathbb{R}^{1000 \times 3}$, $\bm{A}_3 \sim \mathcal{U}(-3,\,2)$, $\bm{B}_3 \sim \mathcal{U}(-0.03,\,0.02)$. $\bm{A}_3$ has sparsity probability of 0.25. 

    For all stages, $\bm{B}_i$ has a sparsity probability of 0.999. The noise matrix $\bm{V}_i \sim \mathcal{N}(0,\,0.001)$. To reproduce the correlations between process variables often observed in practice, we made $\bm{X}_i$ a low-rank matrix with bell-shaped singular values where approximately four singular vectors are required to explain 90\% of the data.
\end{itemize}

We used the same configurations to generate other datasets except the number of process variables. In particular, for Dataset-2,-3, -4, and -5, the number of process variables is multiplied by 2, 5, 10, and 100 respectively.

%
\bibliographystyle{elsarticle-num} 
\bibliography{cas-refs}





\end{document}